%% file: main.tex
\documentclass{article}

\usepackage{microtype}
\usepackage{graphicx}
\usepackage{subfigure}
\usepackage{booktabs} 
\usepackage{hyperref}

\usepackage[noend]{algorithmic}

\usepackage[accepted]{mlsys2025/mlsys2025}

\usepackage{amsmath}

\usepackage{balance}
\usepackage{enumitem}
\usepackage{multirow}
\usepackage{xspace}
\usepackage[compact]{titlesec}
\usepackage{booktabs}
\usepackage{placeins}

\newcommand{\systemname}{Armada\xspace}
\newcommand{\partitioning}{GREM\xspace}
\newcommand{\newparagraph}[1]{{\vspace{2pt}\noindent\textbf{#1}}}

\mlsystitlerunning{}

\begin{document}

\twocolumn[
\mlsystitle{\systemname: Memory-Efficient Distributed Training\\of Large-Scale Graph Neural Networks}

\begin{mlsysauthorlist}
\mlsysauthor{Roger Waleffe}{uw}
\mlsysauthor{Devesh Sarda}{uw}
\mlsysauthor{Jason Mohoney}{uw}
\mlsysauthor{Emmanouil-Vasileios Vlatakis-Gkaragkounis}{uw}
\mlsysauthor{Theodoros Rekatsinas}{apple}
\mlsysauthor{Shivaram Venkataraman}{uw}
\end{mlsysauthorlist}

\mlsysaffiliation{uw}{University of Wisconsin-Madison}
\mlsysaffiliation{apple}{Apple}

\mlsyscorrespondingauthor{Roger Waleffe}{waleffe@wisc.edu}

\vskip 0.3in

\begin{abstract}
\input{sections/s0_abstract}
\end{abstract}
]

\printAffiliationsAndNotice{}

\input{sections/s1_intro}
\input{sections/s2_preliminaries}
\input{sections/s3_overview}
\input{sections/s4_partitioning}
\input{sections/s5_implementation}
\input{sections/s6_eval}
\input{sections/s7_related_work}
\input{sections/s8_conclusion}

\clearpage
\bibliography{ref}
\bibliographystyle{mlsys2025/mlsys2025}

\end{document}

%% file: sections/s0_abstract.tex
We study distributed training of Graph Neural Networks (GNNs) on billion-scale graphs that are partitioned across machines. Efficient training in this setting relies on min-edge-cut partitioning algorithms, which minimize cross-machine communication due to GNN neighborhood sampling. Yet, min-edge-cut partitioning over large graphs remains a challenge: State-of-the-art (SoTA) offline methods (e.g., METIS) are effective, but they require orders of magnitude more memory and runtime than GNN training itself, while computationally efficient algorithms (e.g., streaming greedy approaches) suffer from increased edge cuts. Thus, in this work we introduce Armada, a new end-to-end system for distributed GNN training whose key contribution is \partitioning, a novel min-edge-cut partitioning algorithm that can efficiently scale to large graphs. \partitioning builds on streaming greedy approaches with one key addition: prior vertex assignments are continuously refined during streaming, rather than frozen after an initial greedy selection. Our theoretical analysis and experimental results show that this refinement is critical to minimizing edge cuts and enables \partitioning to reach partition quality comparable to METIS but with 8-65$\times$ less memory and 8-46$\times$ faster. Given a partitioned graph, Armada leverages a new disaggregated architecture for distributed GNN training to further improve efficiency; we find that on common cloud machines, even with zero communication, GNN neighborhood sampling and feature loading bottleneck training. Disaggregation allows Armada to independently allocate resources for these operations and ensure that expensive GPUs remain saturated with computation. We evaluate Armada against SoTA systems for distributed GNN training and find that the disaggregated architecture leads to runtime improvements up to 4.5$\times$ and cost reductions up to 3.1$\times$.

%% file: sections/s1_intro.tex
\section{Introduction}
Graph Neural Networks (GNNs) have emerged as the defacto approach for machine learning over graph-structured inputs~\cite{chami2021machine}; GNN-based models are currently used in navigation apps~\cite{derrow2021eta}, to predict protein structures~\cite{jumper2021highly}, and to create weather forecasts (GraphCast~\cite{lam2022graphcast}). These impressive results, however, require training GNNs over massive amounts of graph data. For example, GraphCast was trained on 53TB over four weeks using 32 Cloud TPU v4 nodes (10/2024 est. cost: \$70K), limiting the development of such a model to those with sufficient resources. 

Motivated by the above, this work focuses on scalable, cost-effective, distributed GNN training over large graphs using common cloud offerings. While recent works~\cite{salient++, distDGL, distdglv2} have sought to address this need, we find that existing pipelines face scalability challenges when graphs have billions of nodes or edges and when training with multiple GPUs. These challenges arise from the unique properties of the GNN workload itself.

In particular, distributed GNN training necessitates that the graph is partitioned across machines; yet, the partitioning has a direct impact on the subsequent training efficiency, as GNN systems must communicate across machines to sample the neighborhood of graph nodes~\cite{shao2024distributed}. This communication can be reduced using \textit{min-edge-cut partitioning} algorithms that minimize the number of edges with endpoints in different partitions (machines) (called \textit{cut edges}). Thus, min-edge-cut partitioning is widely used in GNN systems, and has been shown to lead to an order of magnitude faster training compared to random partitioning~\cite{merkel2023experimental, distdglv2}. 

Min-edge-cut partitioning, however, becomes increasingly expensive with graph size. For instance, many systems utilize the offline algorithm METIS~\cite{karypis1997metis} due to its ability to effectively minimize edge cuts by iteratively refining partitions across the whole graph and its comparatively efficient implementation~\cite{merkel2023experimental, shao2024distributed, lin2023comprehensive}; yet, METIS takes 8000s and requires a special machine with 630GB of memory to partition a common benchmark graph (the 1.6B edge OGBN-Papers100M), whereas GNN training takes only 549s (10 epochs, one GPU) and can run on cloud machines with 244GB of memory~\cite{mariusgnn} (details in Section~\ref{sec:eval}). Although the partitioning overhead can be amortized across models, it still presents a bottleneck to GNN training. To address this issue, streaming algorithms iterate over the graph and assign vertices to partitions greedily~\cite{abbas2018streaming}. While these algorithms offer improved scalability, they tend to result in more edge cuts than offline methods~\cite{zhang2018akin}; e.g., we find a streaming greedy approach cuts up to 4$\times$ more edges than METIS.

In this work, we introduce Armada, a new end-to-end system for large-scale distributed GNN training that aims to address the bottleneck of partitioning in existing GNN pipelines. To overcome this challenge, Armada's key contribution is a novel memory-efficient min-edge-cut partitioning algorithm called \partitioning (Greedy plus Refinement for Edge-cut Minimization). \partitioning can efficiently scale to massive graphs on common hardware by processing streaming chunks of graph edges, yet it still returns partitions with edge cuts comparable to METIS. For example, in the same setting in which METIS requires 8000s and 630GB, \partitioning can partition the graph with similar edge cuts in 175s using 9.3GB.

\partitioning's partitioning algorithm builds on existing streaming greedy approaches. 
Specifically, \partitioning iterates over the graph edges in chunks and greedily assigns the vertices in each chunk to partitions. The key idea behind \partitioning, however, is that it allows prior vertex assignments to be modified throughout the process, rather than freezing them after an initial greedy selection (as in existing algorithms~\cite{abbas2018streaming}). This approach, inspired by offline algorithms, refines the partitioning by leveraging lightweight statistics accumulated during streaming (these statistics provide estimates of the number of neighbors per node in each partition).

We analyze theoretically \partitioning's expected number of edge cuts versus chunk size, providing insight into its expected behavior. This analysis, confirmed by experiments, shows that refinement is critical for minimizing edge cuts when using small chunk sizes (e.g., $\le$10\% of the edges) and thus for minimizing \partitioning's computational requirements (which are proportional to chunk size): We show that \partitioning with a chunk size of 10\% and METIS cut a similar number of edges, but \partitioning does so with 8$\times$ less memory and runtime (see Section~\ref{sec:eval}). \partitioning even achieves comparable results with a chunk size of 1\%, leading to further reductions and enabling \partitioning to partition the largest public graphs (e.g., Hyperlink-2012~\cite{hyperlink}; 3.5B nodes, 128B edges) with only 500GB of memory.

Given a partitioned graph, Armada's second main contribution is the introduction of a new distributed architecture, that disaggregates the CPU resources used for neighborhood sampling from the GPU resources used for model computation, in order to achieve scalable, memory-efficient, and cost-effective GNN training on common hardware. Concretely, Armada consists of: 1) A partitioning layer that implements \partitioning. 2) A storage layer to store the partitioned graph, implemented over cheap disk-based storage. 3) A distributed mini batch preparation layer consisting of a set of workers running on cheap CPU-only machines; workers read graph partitions from storage and prepare batches (i.e., perform neighborhood sampling) for training. 4) A distributed model computation layer that utilizes a set of GPU machines to perform training over the prepared batches.

We chose a disaggregated architecture to optimize resource utilization during training. On common cloud machines, we find that even with zero communication, mini batch preparation can be up to an order of magnitude slower than mini batch computation (Figure~\ref{fig:armada_breakdown}). Disaggregation allows Armada to overcome this imbalance. By independently scaling the batch preparation layer, we can ensure that GPUs in the computation layer remain saturated during training. In contrast, existing systems, which rely only on the fixed set of CPU resources attached to the GPU machines used for training to prepare batches, are unable to parallelize mini batch preparation and suffer from sublinear speedups as compute resources are scaled. For example, on a cloud GPU machine, we find that two SoTA systems~\cite{salient++, distDGL} yield only 4.3$\times$ and 1.7$\times$ speedup when using eight instead of one GPU (Table~\ref{tab:runtime_nc} left). Sublinear speedups lead to higher than necessary total training cost and runtime over massive graphs, as expensive compute resources sit idle. Yet in the same setting, \systemname achieves a 7.5$\times$ speedup with eight instead of one GPU.

Despite the flexibility of disaggregation, challenges arise due to the communication overhead between various components. Thus, we carefully design Armada with a focus on minimizing communication between and within layers. In particular, Armada includes two optimizations to reduce the data sent between batch preparation and compute workers: 1) batch workers group mini batches destined for different GPUs on the same compute worker and transfer them together, rather than independently, in order to enable greater compression (mini batch grouping), and 2) compute workers in Armada maintain a cache of frequently accessed data in their local CPU memory (feature caching). Together, these optimizations enable Armada to scale each layer in the architecture independently without communication bottlenecks.

We evaluate Armada's disaggregated architecture for GNN training and compare against existing SoTA systems. Using popular GNN architectures, we show that while existing systems scale sublinearly, Armada does not, leading to runtime improvements up to 4.5$\times$ and monetary cost reductions up to 3.1$\times$ compared to existing systems.

%% file: sections/s2_preliminaries.tex
\section{Preliminaries}
\label{sec:prelim}
We discuss necessary background on GNN training.

\subsection{GNNs and GNN Mini Batch Training}
\label{subsec:prelim_gnns}
GNNs achieve state-of-the-art (SoTA) accuracy by learning to combine information about graph nodes with information from their multi-hop neighborhood. The local information for each node is encoded in a \textit{feature vector} that can be fixed or learned during training. All feature vectors are stored together in a lookup table indexed by node ID---For large graphs, the storage overhead for this table can require hundreds of GBs to TBs of memory~\cite{mariusgnn}.

As such, large-scale GNN training is typically performed in a mini batch fashion consisting of two distinct parts: \textit{mini batch preparation} and \textit{mini batch computation}. Mini batch preparation is uniquely challenging for GNNs~\cite{chami2021machine, dorylus, p3gnn, salient}; it requires sampling multi-hop neighborhoods~\cite{graphsage} for a batch of nodes and loading the corresponding feature vectors from the graph. Given the storage overhead of the latter (above), mini batch preparation typically occurs on CPUs with access to ample DRAM. After a batch is prepared, mini batch computation (the GNN forward/backward pass) consists of matrix multiplies that call for GPU acceleration. Thus, batches prepared on CPUs are then transferred to GPUs for computation.

\subsection{Distributed GNN Training}
\label{subsec:prelim_distributed}
When the storage overhead of a graph exceeds the CPU memory capacity of a single machine, or when parallelization is desired to accelerate training, prior works propose to use distributed training~\cite{shao2024distributed}. In this case, graph nodes (and their features) are split into disjoint \textit{partitions} that are loaded into separate machines. In this setup, mini batch preparation is also distributed---Each machine is responsible for preparing batches in parallel and sampling the required multi-hop neighborhoods across the whole graph, by communicating with other machines as needed.

\newparagraph{The Need for Scalable Min-Edge-Cut Partitioning}
Cross-machine neighborhood sampling and feature loading can lead to a communication bottleneck that fundamentally limits the scalability and throughput of batch preparation across the set of machines~\cite{salient++}. To mitigate this issue, existing systems rely on partitioning algorithms that minimize the number of cross-partition edges~\cite{distdglv2}. As highlighted in the introduction, however, current partitioning algorithms that minimize edge cuts are expensive---partitioning often dominates the overall GNN runtime and limits the maximum graph size that can be processed given certain resources.

\begin{figure}[t]
  \centering
  \includegraphics[width=.45\textwidth]{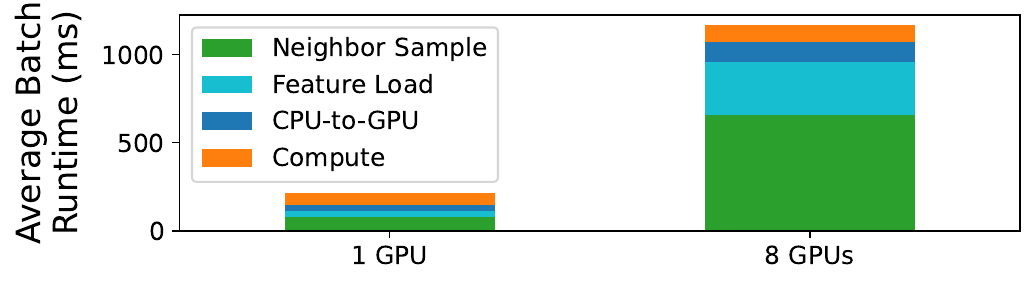}
  \vspace{-0.15in}
  \caption{Breakdown of the average runtime per training iteration in the SoTA system MariusGNN (GraphSage-Large on OGBN-Papers100M; details in Section~\ref{sec:eval}). Neighborhood sampling plus feature loading on the CPU dominates GNN runtime.}
  \label{fig:armada_breakdown}
  \vspace{-0.15in}
\end{figure}

\newparagraph{The Need For Disaggregated Training}
We find that \textit{even when there is zero communication}, mini batch preparation can bottleneck distributed GNN training in existing systems, leading to GPU underutilization and unnecessarily expensive training. This problem is exacerbated on common cloud machines with fast GPUs and fixed CPU resources. 

For example, in Figure~\ref{fig:armada_breakdown} we show the average time for mini batch preparation and computation across training iterations on a common GNN benchmark. Figure~\ref{fig:armada_breakdown} shows that multi-hop sampling (even when optimized~\cite{mariusgnn}) and feature loading---which together encompass mini batch preparation---dominate overall training time. Furthermore, this overhead increases with increasing GPUs (distributed data parallel training with weak scaling, commonly used for GNN training, requires preparing one mini batch per GPU). 

\begin{figure}[t]
  \centering
  \includegraphics[width=.45\textwidth]{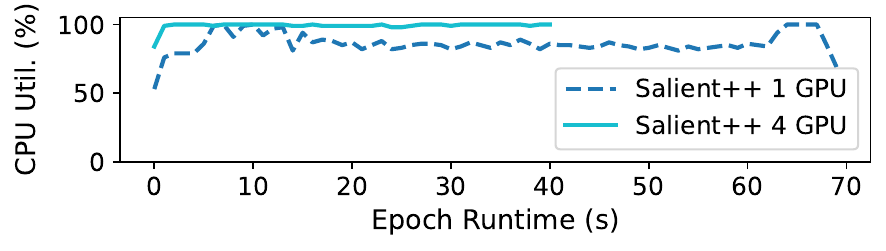}
  \vspace{-0.15in}
  \caption{CPU utilization in the SoTA system Salient++ (details in Section~\ref{sec:eval}; GraphSage-Small on OGBN-Papers100M). Nearly all CPU resources are used to parallelize mini batch preparation and minimize training time with one GPU; the CPU resources are insufficient for multi-GPU training, leading to sublinear speedups.}
  \label{fig:salient_cpu}
  \vspace{-0.15in}
\end{figure}

Given the runtime discrepancy, it's necessary to parallelize mini batch preparation across CPUs in order to keep GPUs busy with computation. Existing systems, however, rely only on the fixed set of CPU resources attached to GPU machines for this parallelization, fundamentally hindering their ability to prepare batches~\cite{salient, distdglv2}. To highlight this issue, in Figure~\ref{fig:salient_cpu}, we show the CPU utilization of the SoTA system Salient++~\cite{salient++} during GNN training. Salient++ requires more than 80 percent of the CPU to prepare batches in parallel for one GPU. When training with four GPUs, the CPU is fully saturated, limiting the throughput of mini batch preparation, leading to sublinear scaling (1.6$\times$ instead of 4$\times$; Table~\ref{tab:runtime_nc}), and resulting in expensive GPUs sitting partially idle.

The above observations motivate a disaggregated system for large-scale GNN training that supports scaling each part of the workload independently. While disaggregation has improved resource efficiency in traditional ML settings~\cite{graur2022cachew, jin2024efficient}, prior work on disaggregated GNN training (Dorylus~\cite{dorylus}) has focused only on utilizing serverless functions and full multi-hop neighborhoods (i.e., no sampling). Full neighborhoods, however, lead to expensive communication for multi-layer GNNs and a serverless architecture limits the type of models that can be trained efficiently without GPUs. On a common GNN benchmark, we find that throughput in Dorylus plateaus at 89.6s/epoch (\$3.75/epoch); GPU-based systems can be 12$\times$ faster and 53$\times$ cheaper (see Table~\ref{tab:runtime_nc} left).

%% file: sections/s3_overview.tex
\section{\systemname: Architecture Overview}
\label{sec:overview}

\begin{figure*}[!t]
  \centering
  \includegraphics[width=1.0\textwidth]{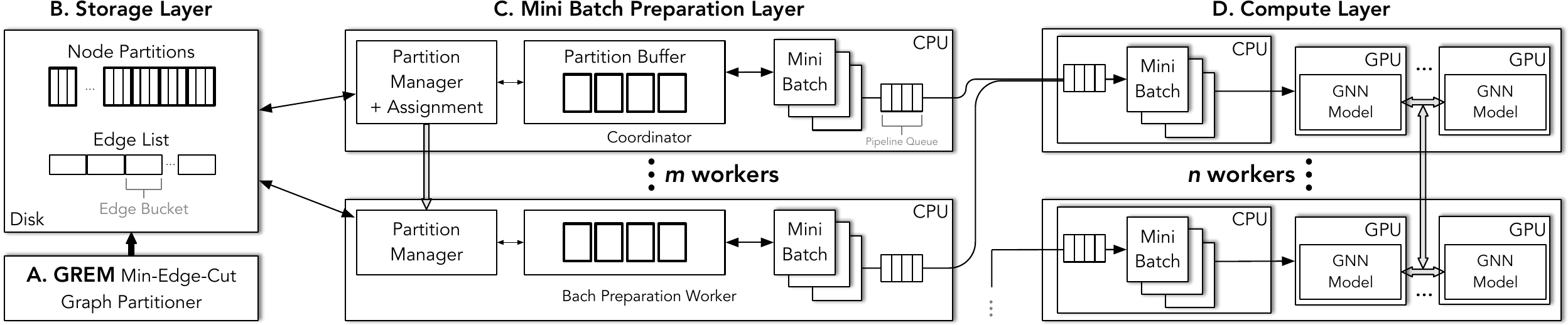}
  \vspace{-0.25in}
  \caption{\systemname system diagram. \textbf{A.} Graph data is partitioned using \partitioning (Section~\ref{sec:partitioning}) and then \textbf{B.} stored on disk in the storage layer. \textbf{C.} A disaggregated mini batch preparation layer loads graph partitions into memory and prepares mini batches for workers in the compute layer. \textbf{D.} The compute workers process these mini batches on GPUs and periodically synchronize dense model parameters.}
  \label{fig:system_diagram}
\end{figure*}

Armada addresses the aforementioned challenges of large-scale distributed GNN training (Section~\ref{sec:prelim}) by introducing a min-edge-cut partitioning algorithm (\partitioning) and by employing a disaggregated architecture (Figure~\ref{fig:system_diagram}). Concretely, Armada consists of four components; we next discuss the responsibilities of each during training, before describing \partitioning in Section~\ref{sec:partitioning} and optimizations to reduce communication across the architecture in Section~\ref{sec:implementation}.

\newparagraph{\partitioning Partitioning Layer}
Given an input graph (stored on disk) in the form of an edge list and a set of features for each node, Armada partitions the nodes into a set of $p$ partitions using \textit{\partitioning}. \partitioning returns a label for each node specifying its partition, and saves this information to disk.

\newparagraph{Storage Layer}
The storage layer in Armada can store the partitioned graph using a variety of common backends (e.g., AWS S3, EBS, HDFS). We store the feature vectors for each node in a partition sequentially and group the graph edges into buckets: For a pair of node partitions $(i, j)$, edge bucket $(i, j)$ contains all edges from nodes in $i$ to nodes in $j$. All edges in each bucket are then stored sequentially as a list. This format allows sets of partitions and the edges between them to be accessed using only sequential file reads/writes.

\newparagraph{Mini Batch Preparation Layer}
Armada uses a distributed set of workers running on cheap CPU-only machines to prepare mini batches for training.
Each worker reads a set of partitions (and the edges between them) from the storage layer into memory. The specific partition assignment for each machine is made by a designated worker, called the \textit{coordinator}, according to a \textit{randomized algorithm} (Section~\ref{sec:implementation}). After loading their assigned partitions, workers construct batches for training. Armada supports both 1) local construction, where machines prepare batches using only the data in their own CPU memory, leading to zero communication across machines, and 2) distributed construction, where multi-hop neighborhoods are sampled across the whole graph in the aggregate CPU memory of the layer. To minimize communication between workers in the latter setting, Armada relies on the min-edge-cut partitioning returned by \partitioning and supports replicating high degree nodes on each worker (Section~\ref{sec:implementation}). Once batches are prepared, each worker pushes them to a specified (when configuring Armada) worker in the compute layer. To minimize the data transferred to the compute workers for each batch, Armada uses \textit{mini batch grouping}---batches destined for different GPUs on the same machine are grouped together for transmission to enable greater data compression (Section~\ref{sec:implementation}).

Because the mini batch preparation layer is disaggregated, the number of workers can be chosen independently from the other layers in Armada. In particular, for workloads bottlenecked by mini batch preparation, Armada can allocate enough CPU resources to ensure that all compute workers (GPUs) remain saturated with computation. Additionally, for massive graphs, Armada can rely on the storage layer for primary graph storage, rather than on the CPU memory of the batch preparation (or compute) layer, providing the option for lower cost, memory-efficient training deployments.

\newparagraph{Compute Layer}
Armada's compute layer consists of a set of machines with attached GPU(s) and is responsible for model computation. Compute workers listen for mini batches from specified worker(s) in the mini batch preparation layer and then perform the GNN forward/backward pass on received batches in parallel. To minimize the amount of data sent for each mini batch, compute workers in Armada also maintain a \textit{feature cache} of frequently accessed features in their local CPU memory (Section~\ref{sec:implementation}). GNN model parameters are replicated across GPUs and model gradients are synchronized before each parameter update. If applicable, gradients for learnable feature vectors are transferred to the CPU memory of the compute worker and then sent back to the corresponding batch worker so it can update its partitions in memory.

%% file: sections/s4_partitioning.tex
\section{Scalable Graph Partitioning}
\label{sec:partitioning}

We now introduce \partitioning, a novel algorithm that enables efficient min-edge-cut graph partitioning over massive graphs on a commodity machine. We describe the optimization objective and algorithm, then analyze it theoretically below.

\subsection{\partitioning: Min-Edge-Cut Partitioning}
\newparagraph{Optimization Objective}
Given a graph $G = (V, E)$, we assume as input an edge list ($E$) stored on disk in a random order. Our goal is to partition the nodes $V$ into a set of $p$ partitions, each of size $\lceil V/p \rceil$ (i.e., a balanced partitioning), according to an algorithm that 1) minimizes the number of cross-partition edges and 2) can scale to massive graphs given a fixed amount of CPU memory (but unlimited disk space) (the edge list for large graphs may not fit in memory on a single, or even multiple machines (e.g., Hyperlink-2012's 128B edges require 2TB~\cite{hyperlink}).

The above problem (balanced min-edge-cut partitioning) is NP-Hard, even for $p=2$; for $p > 2$, no finite approximation algorithm exists unless P=NP~\cite{andreev2004balanced}. As such, existing algorithms rely on heuristics: As highlighted in the introduction, offline algorithms (e.g., METIS) operate over the whole graph and effectively minimize edge cuts through iterative partition refinement, but they face scalability challenges. On the other hand, streaming greedy approaches~\cite{stanton2014streaming, alistarh2015streaming, patwary2019window, stanton2012streaming, faraj2022buffered, petroni2015hdrf, jain1998greedy, tsourakakis2014fennel} have better scalability, but often lead to partitionings with more edge cuts due to their use of fixed greedy decisions (e.g., 4$\times$ more edge cuts than METIS).

\newparagraph{Key Idea}
To combine the advantages of offline and streaming methods, \partitioning employs a streaming greedy approach, but with one key addition: Rather than freezing the partition assignment for a node after an initial greedy selection, \partitioning leverages running statistics accumulated during streaming to continuously reevaluate prior assignments and refine the result. Inspired by offline algorithms, this refinement is critical to minimizing edge cuts.

\begin{algorithm}[t]\small
    \caption{\partitioning Bipartite Graph Partitioning}
    \label{alg:partition_algo} 
    \begin{algorithmic}[1] 
        \REQUIRE num\_nodes; edges; c\_size: chunk size; \texttt{seed\_algo}: seed partition algorithm; P: max partition size (in nodes)
    
        \STATE parts = \texttt{minus\_ones}(num\_nodes); $\;$ sizes = \texttt{zeros}(2) 
        
        \STATE nbr\_counts = \texttt{zeros}(num\_nodes, 2)

        \FOR{$i = 0$ \TO $\texttt{ceil}(\texttt{len}(\text{edges})/\text{c\_size}) - 1$}
            \STATE c\_edges = \texttt{read}(edges[i $*$ c\_size : (i + 1) $*$ c\_size])
            \STATE c\_nodes = \texttt{unique\_nodes\_in\_edges}(c\_edges)

            \IF{$i == 0$}
                \STATE parts[c\_nodes] = \texttt{seed\_algo}(c\_edges, num=2)
                \STATE sizes = [\texttt{num\_zeros}(parts), \texttt{num\_ones}(parts)]
                
                \STATE $\text{nbrs}_0$, nbrs$_1$ = \texttt{cnt\_nbrs}(c\_nodes,$\,$c\_edges,$\,$parts)
                
                \STATE nbr\_counts[c\_nodes] = [$\text{nbrs}_0$, nbrs$_1$]
            \ELSE
                \FOR{n $\in$ c\_nodes}
                    \STATE old\_part = parts[n]
                
                    \STATE $\text{nbrs}_0$, nbrs$_1$ = \texttt{cnt\_nbrs}(n, c\_edges, parts)
                    
                    \IF{old\_part $\neq -1$}
                        \STATE $\text{nbrs}_0$, $\text{nbrs}_1$ = (nbr\_counts[n] + [$\text{nbrs}_0$, $\text{nbrs}_1$]) / 2
                    \ENDIF
                    
                    \STATE parts[n] = \texttt{assign}($\text{nbrs}_0$, $\text{nbrs}_1$, sizes, P)

                    \STATE sizes = \texttt{fix\_sizes}(parts[n], old\_part, sizes)
                    
                    \STATE nbr\_counts[n] = [$\text{nbrs}_0$, nbrs$_1$]
                \ENDFOR
            \ENDIF
        \ENDFOR
    \end{algorithmic}
\end{algorithm}

\newparagraph{Detailed Algorithm}
\partitioning partitions an input graph into $p=2$ partitions as described in Algorithm~\ref{alg:partition_algo}; we focus on $p=2$ because \partitioning returns a partitioning for $p>2$ by first partitioning the graph into two parts, and then recursively re-partitioning each part into two new parts as needed. We identify the two partitions by index \textit{zero} and \textit{one}. Each node starts unassigned (index \textit{minus one}) and each partition starts with size zero (Line 1). We also initialize two numerical values for each node (\texttt{nbr$\_$counts}, Line 2); the purpose of these values is to provide a running estimate of the number of neighbors each node has in each partition. 

\partitioning then proceeds by iterating over the edge list in chunks (Line 3). For each chunk, the edges are loaded into memory (\texttt{c$\_$edges}) and the set of unique nodes (\texttt{c$\_$nodes}) contained in those edges is computed (Lines 4-5). For the first chunk (Line 6), as there are no existing partition assignments that can be used to make greedy decisions, we use a \textit{seed} partitioning algorithm on the in-memory edges (e.g., METIS) to assign all nodes in memory to one of the two partitions (Line 7). The partition sizes and the estimated number of neighbors per node in each partition (calculated based on the in-memory edges and existing partitioning; Algorithm~\ref{alg:partition_algo_helpers} - \texttt{cnt\_nbrs}) are then updated (Lines 8-10). 

For the remaining chunks (Line 11), \partitioning assigns nodes to partitions greedily. For each node $n$ (Line 12), we start by estimating the number of neighbors in each partition using the current chunk's edges and most recent partition assignments (Line 14). If the node is unassigned (i.e., this is the first chunk containing the node), these neighbor estimates are used directly: To minimize edge cuts, we assign the node to the partition containing most of its neighbors, unless the partition is full (Line 17; Algorithm~\ref{alg:partition_algo_helpers} - \texttt{assign}). The partition sizes are then updated (Line 18; Algorithm~\ref{alg:partition_algo_helpers} - \texttt{fix\_sizes}) and the neighbor estimates for the node are saved (Line 19). \textit{The algorithm, as described so far, represents a streaming greedy approach with fixed assignments.} 

Instead of fixing an initial greedy decision for each node, \partitioning reevaluates a node's partition assignment each time it reappears in memory. Specifically, for a previously assigned node $n$ (Line 15), we refresh our estimate of the number of neighbors in each partition using an average of the estimate from the current chunk and the estimates accumulated from prior chunks (Lines 16). Node $n$ is then assigned to a partition greedily using these updated estimates (Line 17), which are then saved for future use (Line 19). We highlight that, by repeatedly averaging the accumulated neighbor estimates with the most recent ones, we are computing a weighted average of the estimates across all prior chunks containing the node, with the weight of each preceding chunk decreasing by a factor of two.

Updating prior greedy assignments based on the weighted average of neighbor estimates has the following advantages: First, nodes (which reappear) are not greedily assigned based on the estimates from only one chunk (as in existing algorithms)---these estimates can be noisy, particularly for small chunks when nodes have only a few neighbors in memory. Second, by weighting the average, more value is placed on recent estimates which are likely to be more accurate (as partition assignments may have changed since prior estimates were computed). The end result is a continuous refinement of greedy decisions throughout the algorithm.

\begin{algorithm}[t]\small
    \caption{\partitioning Helper Functions}
    \label{alg:partition_algo_helpers}
    
    \begin{algorithmic}[1]

    \STATE \texttt{cnt\_nbrs}(nodes, edges, parts):
    \begin{ALC@g}
        \STATE local\_nbr\_counts = \texttt{zeros}(\texttt{len}(nodes), 2)
        \FOR{n $\in$ nodes}
            \FOR{(src, dst) $\in$ edges}
                \IF{src $==$ n \OR dst $==$ n}
                    \STATE nbr = src \textbf{if} dst $==$ n \textbf{else} dst
                    
                    \IF{parts[nbr] $\neq$ -1}
                        \STATE local\_nbr\_counts[n][parts[nbr]] += 1
                    \ENDIF
   
                \ENDIF
            \ENDFOR
        \ENDFOR
        \RETURN local\_nbr\_counts
    \end{ALC@g}

    \STATE \texttt{assign}($\text{nbrs}_0$, $\text{nbrs}_1$, sizes, P):
    \begin{ALC@g}
        \STATE \textbf{if} $\text{nbrs}_0$ $<$ $\text{nbrs}_1$ \AND sizes[1] $<$ P \textbf{then} \textbf{return} 1

        \STATE \textbf{if} $\text{nbrs}_1$ $<$ $\text{nbrs}_0$ \AND sizes[0] $<$ P \textbf{then} \textbf{return} 0

        \RETURN \texttt{arg\_min}(sizes)
        
    \end{ALC@g}

    \STATE \texttt{fix\_sizes}(new\_part, old\_part, sizes):
    \begin{ALC@g}
        \STATE sizes[new\_part] += 1
        \STATE \textbf{if} old\_part $\neq -1$ \textbf{then} sizes[old\_part] $-$= 1
    \end{ALC@g}
    
    \end{algorithmic}
    
\end{algorithm}

\subsection{Theoretical Analysis of \partitioning}
\label{subsec:partitioning_analysis}

We now analyze the number of edge cuts returned by \partitioning versus chunk size. We focus on chunk size as it directly affects the computational overhead of the algorithm. As chunk size decreases, so does \partitioning's memory requirement and runtime; only the active chunk of edges needs to be in memory and the time for the initial seed partitioning algorithm on the first chunk dominates the time for the simple greedy processing of subsequent chunks. We compare the expected number of edge cuts when using fixed greedy assignments to that of the refined greedy assignments employed by \partitioning.

\newparagraph{Fixed Greedy Assignments}
We focus on the assignment of a specific node $n$ and assume all other nodes are assigned to partitions. Among all edges, let node $n$ have $k$ neighbors, with $k_0$ in partition zero, and $k_1$ in partition one. Without loss of generality, we assume $k_0 \ge k_1$. Observe that, with a chunk size of $|E|$ (i.e., all edges), our greedy algorithm will assign node $n$ to partition zero to minimize edge cuts. 

To analyze the effect of chunk size, we ask, what is the probability node $n$ will be assigned to partition zero if only $|E|*x$ edges (sampled uniformly) are used to make the decision (i.e., if we use a chunk size of $|E|*x$)? Let $k'_0$ and $k'_1$ be the number of neighbors of node $n$ in partition zero and one that are present in the sampled $|E|*x$ edges. Then we seek to calculate $Pr(k'_0 \ge k'_1 | k_0 \ge k_1, x)$. We assume that $k'_0 + k'_1 = x*k$ (i.e., sampling $|E|*x$ edges leads to sampling $k*x$ neighbors). Then $k'_0$ (or $k'_1$) is a random variable sampled from a Hypergeometric distribution describing the probability of sampling (without replacement) a specific number of neighbors in partition zero (one) from a finite population of size $k$, containing $k_0$ ($k_1$) total neighbors in partition zero (one), using $k*x$ draws. We also have that $k'_1 = k*x - k'_0$ and $Pr(k'_0 \ge k'_1) = Pr(k'_0 \ge k*x - k'_0) = Pr(k'_0 \ge 0.5*k*x) = 1 - Pr(k'_0 < 0.5*k*x)$. The latter can be calculated using the cumulative distribution function (CDF) of the Hypergeometric distribution and describes the probability of correctly assigning node $n$ given a chunk size of $|E|*x$ (correct here means making the same greedy decision as the one made if all edges are available).

Given the probability of correctly assigning node $n$, we can calculate the expected number of correctly assigned nodes $T$ in the whole graph. Assuming nodes are independent, we have:
$E[T] = \sum_{i=1}^{|V|}(1 - Pr(k^{i'}_0 < 0.5*k^i*x))$ 
with $k^{i'}_0 \sim$ Hypergeometric($k^i$, $k^i_0$, $x$), $k^i$ the number of neighbors (among all edges) of node $i$, and $k^i_0$ the number of these neighbors in the partition containing more of node $i$'s neighbors. Finally, the expected number of edge cuts $C$ is:
\begin{align}
    \label{eqn:greedy_edge_cuts}
    E[C] = \sum_{i=1}^{|V|}&(k^i - k^i_0)*(1 - Pr(k^{i'}_0 < 0.5*k^i*x))\\ +& k^i_0*Pr(k^{i'}_0 < 0.5*k^i*x) \nonumber
\end{align}
since $k^i - k^i_0$ edges are cut for node $i$ if it is correctly assigned and $k^i_0$ edges are cut otherwise. Equation~\ref{eqn:greedy_edge_cuts} can be calculated given $k^i$ and $k^i_0$ for each node $i$ ($k^i_0$ can be estimated given an existing graph partitioning or by making assumptions about a graph's connectivity).

\newparagraph{The Benefit of Refinement}
We now ask how the expected number of edge cuts $E[C]$ changes if greedy decisions are updated (refined) based on a weighted average of neighbor estimates across chunks (as in \partitioning). We focus on the simplest case: We assume two chunks ($\alpha$ and $\beta$), each of size $|E|*x$ are used to assign a given node $n$ to a partition. Let $k'_{0, \alpha}$ and $k'_{0, \beta}$ be the number of neighbors of node $n$ in partition zero among the sampled edges in chunk $\alpha$ and $\beta$ respectively (and likewise for $k'_{1, \alpha}$, $k'_{1, \beta}$ and partition one). In the two chunk case, the weighted average simplifies to a regular average (which can be simplified to a sum): We seek to calculate $Pr(k'_{0, \alpha} + k'_{0, \beta} \ge k'_{1, \alpha} + k'_{1, \beta} | k_0 \ge k_1, x)$.

Observe that $k'_{0, \alpha} + k'_{0, \beta}$ is the number neighbors of node $n$ in the $2*(|E|*x)$ edges formed by the union of chunk $\alpha$ and $\beta$ (each chunk is disjoint). Given this, the expected number of cut edges $E[C]$, when averaging over two chunks each of size $|E|*x$, can be calculated using Equation~\ref{eqn:greedy_edge_cuts} with $x$ replaced by $2x$. In other words, refinement across chunks increases the \textit{effective chunk size} (but not actual chunk size) of the algorithm, leading to better neighbor estimates. Similar intuition applies when generalizing the analysis beyond two chunks, which we omit for brevity.

In Figure~\ref{fig:grem_ablation}, based on the analysis in this section, we plot the expected number of edge cuts $E[C]$ versus chunk size with and without refinement. Figure~\ref{fig:grem_ablation} highlights that refining greedy assignments based on neighbor estimates averaged across multiple chunks leads to fewer edge cuts, particularly for small chunk sizes; in fact, with this refinement, \partitioning can partition the graph with near minimal edge cuts even with chunk sizes $\le$10\%. See Section~\ref{subsec:eval_partitioning} for more details. 

%% file: sections/s5_implementation.tex
\section{Disaggregated Implementation}
\label{sec:implementation}
Given a min-edge-cut partitioned graph, \systemname employs a disaggregated architecture to enable cost-effective, distributed GNN training (Section~\ref{sec:overview}). We now describe important design details that allow \systemname to independently scale each layer and minimize communication in this architecture.

\newparagraph{Partition Assignment to Batch Construction Workers}
We first discuss the randomized algorithm used by Armada's designated worker, called the \textit{coordinator}, to assign partitions in the storage layer to machines in the mini batch preparation layer in order to complete one round of training. To support scaling each of these two layers independently, we require that the algorithm has the following guarantee: all partitions (and thus graph nodes used for training) must appear in memory at least once per epoch, regardless of whether the full graph (all partitions) fits in the aggregate CPU memory of the batch construction workers or not\footnote{If the full graph does not fit in the aggregate CPU memory of the batch construction workers, then neighborhood sampling cannot be done over the full graph, but can instead be done over the entire subgraph in the aggregate CPU memory of the layer.}.

The coordinator assigns partitions to workers as follows: First, the partitions are randomly split into disjoint subsets, one for each worker. This split occurs without data movement (only a mapping is maintained on the coordinator). Given an assigned set of partitions, each worker can then begin training by loading as many random partitions from its subset into memory as possible. After processing these partitions, any remaining partitions assigned to the worker are swapped in for training one by one in a random order. 

Randomized assignment, while simple, satisfies two desired properties: 1) opportunities for parallelism are maximized, as each batch worker operates on a disjoint set of partitions, and 2) each partition is read from the storage layer exactly once, ensuring all nodes appear in memory with minimal I/O between the two layers. \systemname, however, can easily support other partition assignment policies. In particular, to further minimize communication between workers due to cross-machine neighborhood sampling (in addition to min-edge-cut partitioning), \systemname supports partial or even entire (memory permitting) feature replication across workers, as done in prior work~\cite{cao2023communication, salient++}. In this case, the nodes to be replicated are placed in a special partition that is assigned to, and kept in memory, on \textit{all} workers. Min-edge-cut partitioning and randomized partition assignment are then used on the remaining nodes.

\newparagraph{Mini Batch Grouping}
We next discuss mini batch grouping, the first of two techniques used by \systemname to minimize data transfer between batch preparation and compute workers.

Mini batch grouping applies when a batch construction worker is responsible for sending data to a compute worker that contains multiple GPUs. In this case, the batch construction worker must prepare and transfer one mini batch per GPU for each training iteration (such that each GPU can process a batch in parallel). \systemname groups these mini batches into a global batch, as mini batches contained in a global batch may require the same nodes. This allows \systemname to optimize feature loading and transfer: \systemname loads and transfers the feature vectors for the unique nodes in a global batch only once and copies them between GPUs as needed.
For compute workers with 8 GPUs, we find mini batch grouping reduces batch preparation time by 1.13$\times$ and transfer time by 1.72$\times$, increasing overall throughput by 1.15$\times$ on the common OGBN-Papers100M graph used in the experiments (Section~\ref{sec:eval}).

\newparagraph{Compute Worker Feature Caching}
Finally, \systemname can further minimize communication between the batch preparation and compute layers by caching feature vectors for frequently accessed nodes locally on compute workers (in CPU memory). In this case, \systemname needs to send only the non-cached features for each (global) batch between layers. Mini batches are then augmented as needed with the additional feature vectors once they are received in CPU memory by compute workers, before being transferred to the GPU(s) for training. We keep batch construction workers informed of the cache contents by listening to and acknowledging messages from the compute workers that describe planned updates and we use a simple LRU caching policy.

%% file: sections/s6_eval.tex
\section{Evaluation}
\label{sec:eval}
We evaluate \partitioning and \systemname's disaggregated architecture on common large-scale graphs and compare against METIS~\cite{karypis1997metis} and the popular SoTA GNN systems DGL (version 2.4)~\cite{dgl, distDGL}, Salient++~\cite{salient++}, and MariusGNN~\cite{mariusgnn}. Our experiments show that:

\begin{enumerate}[topsep=0pt, left=0pt, noitemsep]
    \item \partitioning can efficiently scale min-edge-cut partitioning to large graphs, leading to up to 45$\times$ and 68$\times$ reduction in runtime and memory overheads compared to METIS.
    \item Disaggregation allows \systemname to achieve scalable, cost-effective GNN training---\systemname achieves a 7.5$\times$ speedup when using eight instead of one GPU when existing SoTA systems yield 4.3$\times$ speedup at best.
\end{enumerate}

\subsection{Experimental Setup}
We start by discussing the setup used in our experiments.

\newparagraph{\systemname and Baseline Details}
MariusGNN supports only single-GPU training, thus we modify it to support multi-GPU training using a standard distributed data parallel architecture. We report results for two versions of \systemname: 1) \systemname and 2) Armada - Aggregated. The former uses the disaggregated architecture described throughout the paper, while the latter uses the CPUs on the GPU machine(s) used for training to prepare batches. The two versions allow us to directly evaluate the benefit of disaggregation. For partitioning, we use \partitioning with Armada and place one partition on each batch construction worker; for baselines, we use their default partitioning plus feature replication and caching (which are METIS-based).

\newparagraph{Hardware Setup}
We partition and train all systems using AWS machines. 
To measure scalability, we use p3.16xlarge instances with eight NVIDIA V100 GPUs and vary how many GPUs are available to each system. These machines contain 64 vCPUs, 488 GiB of CPU memory, and 128 GiB of aggregate GPU memory. For \systemname, we use additional m6a.16xlarge machines for mini batch preparation. These machines have 64 vCPUs and 256 GiB of CPU memory. 

\newparagraph{Datasets, Models, and Metrics}
We report results using Open Graph Benchmark (OGB) datasets~\cite{hu2020ogb, hu2021ogblsc}; we use OGBN-Papers100M (111M nodes, 1.6B edges) and OGB-WikiKG90Mv2 (91M nodes, 601M edges) for large-scale studies, and OGBN-Products (2.5M nodes, 62M edges) plus FB15K-237~\cite{fb15k237} (14.5K nodes, 272K edges) for microbenchmarks. We train a three-layer GraphSage GNN on these datasets with two different hidden sizes: 256 (\textit{GraphSage-Small}) and 1024 (\textit{GraphSage-Large}). The former allows us to run experiments using a data-bound model while the latter aims to represent a compute-bound model. In both cases, we use 30, 20, and 10 neighbors per layer sampled from both incoming and outgoing edges, as done in~\cite{mariusgnn}. For partitioning experiments, we measure the resulting number of edge cuts, runtime, and peak memory usage. For GNN training, we run for 10 epochs and measure runtime and monetary cost. We do not include the time to partition when reporting GNN training times (we report partitioning time independently). We average experiments over three runs.

\newparagraph{Hyperparameters}
We use the same hyperparameters for GNN model architecture and training across systems (e.g., model hidden dimension, number of neighbors, batch size, etc.). These hyperparameters are chosen based on values from prior works~\cite{hu2020ogb, mariusgnn}. For hyperparameters specific to the throughput of each system (e.g., the number neighborhood sampling workers), we manually tune them and select the best configuration.

\subsection{Evaluating \partitioning Partitioning}
\label{subsec:eval_partitioning}
We now evaluate \partitioning and compare to METIS, the SoTA min-edge-cut algorithm used by existing GNN systems.

\begin{figure}[t]
  \centering
  \includegraphics[width=.4\textwidth]{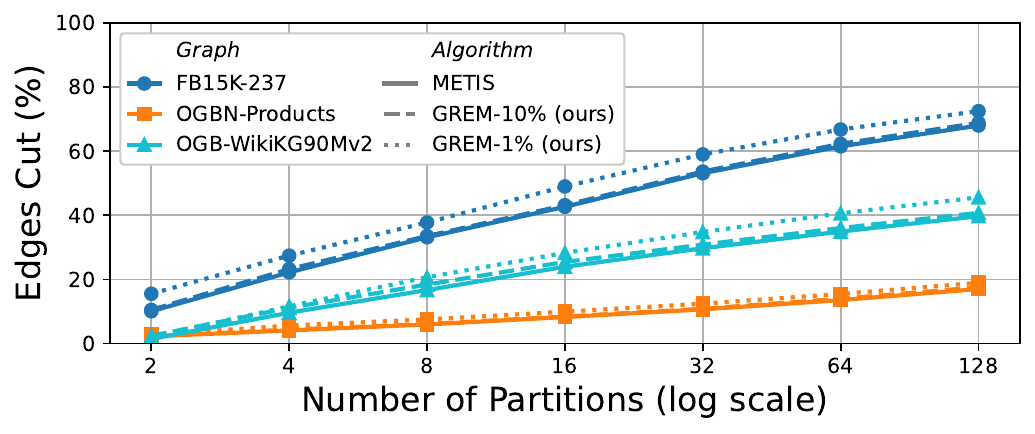}
  \vspace{-0.15in}
  \caption{Percentage of edges cut when using \partitioning versus METIS on three common graphs. \partitioning  achieves comparable edge cuts to METIS, even with a chunk size of just 10\% or 1\%.}
  \label{fig:grem_acc}
  \vspace{-0.1in}
\end{figure}

\newparagraph{Partitioning Quality: Number of Edge Cuts}
In Figure~\ref{fig:grem_acc}, we show the number of edge cuts that result from running \partitioning and METIS on three common graphs. With a chunk size of 10\%, \partitioning partitions the graph with similar quality to METIS. For example, in the most challenging case ($p=128$ partitions), \partitioning cuts just 0.5\% and 1\% more of the graph than METIS on OGBN-Products and OGB-WikiKG90Mv2 respectfully. \partitioning even achieves comparable results with a chunk size of 1\%. Overall, Figure~\ref{fig:grem_acc} shows that with small chunk sizes (e.g., $\le 10$\%), \partitioning can partition graphs with comparable edge cuts to METIS.

\begin{figure}[t]
  \centering
  \includegraphics[width=.48\textwidth]{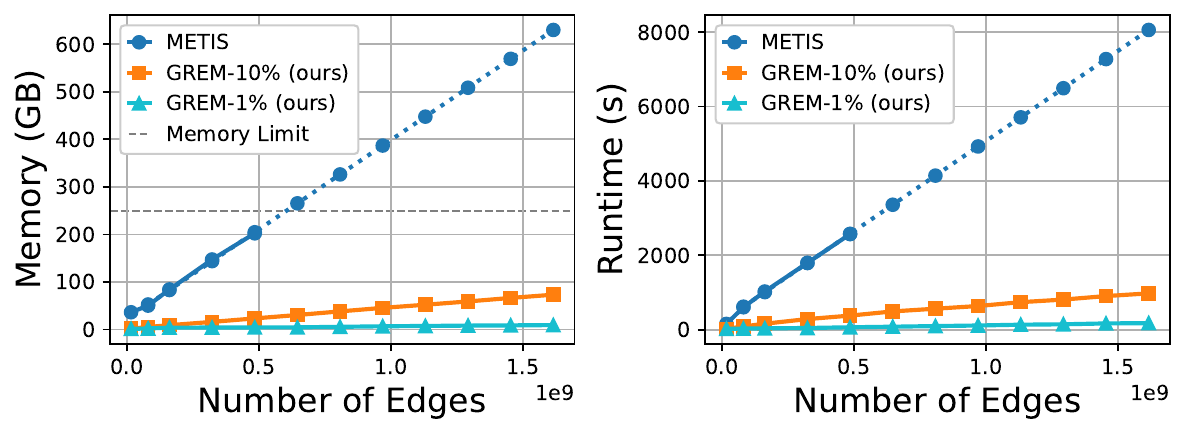}
  \vspace{-0.3in}
  \caption{Memory usage and runtime of \partitioning and METIS when partitioning subgraphs of OGBN-Papers100M of various size. \partitioning reduces the computational requirements of partitioning.}
  \label{fig:grem_memory_runtime}
  \vspace{-0.15in}
\end{figure}

\newparagraph{Partitioning Overhead: Runtime and Memory}
Next, we evaluate the peak memory usage and runtime of \partitioning versus METIS. To do so, we use both algorithms to partition subgraphs of varying size (number of edges), taken from OGBN-Papers100M (1.6B edges total), into two parts ($p=2$). Results are shown in Figure~\ref{fig:grem_memory_runtime}. We plot only $p=2$ for simplicity; as $p$ increases, peak memory remains constant and the runtime of each algorithm increases by the same factor (both \partitioning and METIS partition recursively for $p>2$). Figure~\ref{fig:grem_memory_runtime} (left) shows that METIS is able to partition 600M edges on the machine used for these experiments (250GB of memory). Based on the scaling of memory and runtime, we estimate that METIS needs 8000s and requires a machine with 630GB of memory to partition the entire OGBN-Papers100M graph; we confirmed this estimate on a special machine with 750GB. \partitioning, however, can partition the entire graph in just 976s with 73GB (8.2 and 8.3$\times$ reduction) or 175s with 9.3GB (46 and 65$\times$ reduction) when using a chunk size of 10\% or 1\% respectively.

\begin{figure}[t]
  \centering
  \includegraphics[width=.4\textwidth]{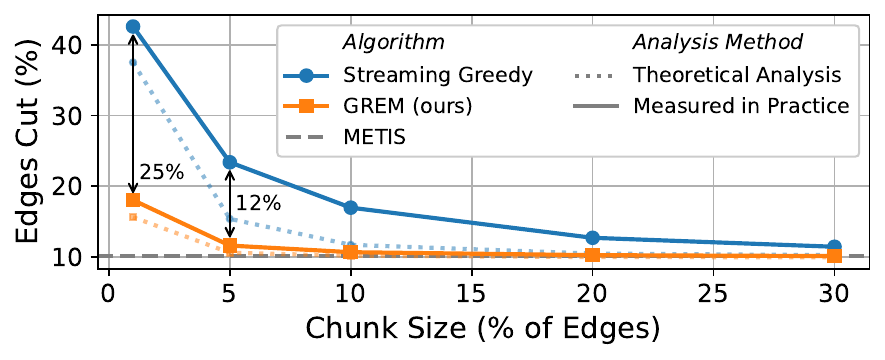}
  \vspace{-0.15in}
  \caption{Percentage of edges cut versus chunk size when using \partitioning on FB15K-237 (the hardest graph to partition in Figure~\ref{fig:grem_acc}) compared to standard streaming greedy approaches. We focus on chunk sizes $\le 30\%$ where the computational benefit of these methods compared to METIS (which is shown for reference, but partitions using the full graph, rather than in chunks) is maximal.}
  \label{fig:grem_ablation}
  \vspace{-0.15in}
\end{figure}

\newparagraph{The Benefit of Refinement}
Finally, we study the benefit of the refined greedy assignments used by \partitioning compared to the fixed greedy assignments of conventional streaming algorithms. For both approaches, we show in Figure~\ref{fig:grem_ablation} the number of edge cuts versus chunk size when partitioning FB15K-237 (the hardest graph to partition in Figure~\ref{fig:grem_acc}) into $p=2$ partitions (given the recursive nature of GREM, similar results hold for $p>2$). We include both the expected number of edge cuts from the theoretical analysis in Section~\ref{subsec:partitioning_analysis}, and the number of edge cuts measured in practice.

Figure~\ref{fig:grem_ablation} shows that for small chunk sizes (e.g., $\le$10\%), refinement is critical to minimizing edge cuts; we observe a reduction of up to 25\% of the graph (at a chunk size of 1\%). These improvements allow GREM to use smaller chunk sizes (e.g., 1-10\%) without suffering a significant increase in edge cuts compared to METIS. For example, with a chunk size of 5\%, \partitioning and METIS differ in edge cuts by $<$1\% of the graph; this difference would be 13\% with fixed greedy assignments. The consequence of these additional edge cuts is slower and more expensive GNN training---We observe that training in Armada is up to 2.4$\times$ slower when using the streaming greedy algorithm rather than \partitioning (for a chunk size of 1\%). This confirms recent results which highlight that high quality partitioning algorithms (e.g., METIS), can lead to faster GNN training compared to streaming greedy approaches (e.g., LDG)~\cite{merkel2023experimental}.

\newparagraph{Summary}
\partitioning can partition large-scale graphs with comparable quality to METIS but with orders of magnitude less computational resources, helping to address the bottleneck of min-edge-cut partitioning for distributed GNN training.

\begin{figure}[t]
  \centering
  \includegraphics[width=0.4\textwidth]{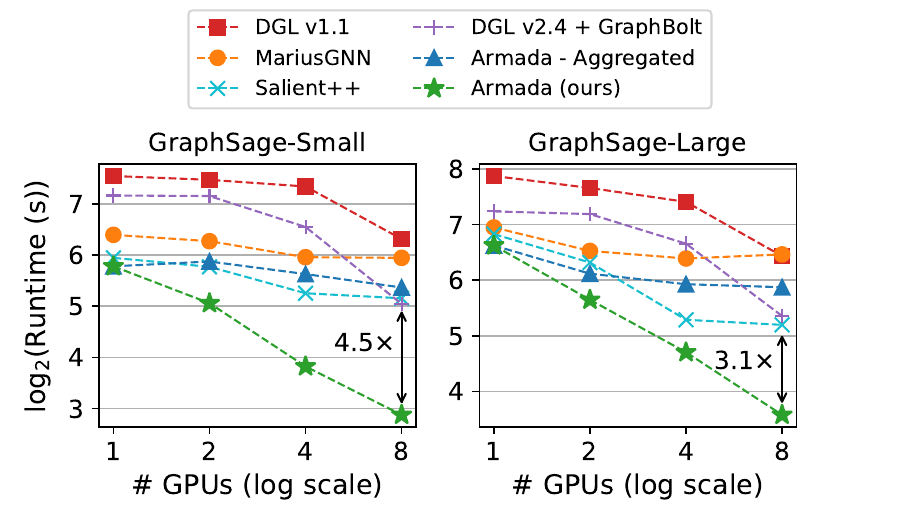}
  \vspace{-0.15in}
  \caption{Epoch runtime versus number of GPUs for DGL, MariusGNN, Salient++, and \systemname using two different GraphSage GNNs on the OGBN-Papers100M dataset. Disaggregation allows \systemname to scale linearly with respect to the number of GPUs.}
  \vspace{-0.1in}
  \label{fig:scaling}
\end{figure}

\subsection{GNN Training: System Comparisons}
Given a partitioned graph, we now evaluate Armada's disaggregated architecture for GNN training. Runtime and cost per epoch for two models on OGBN-Papers100M with \systemname and existing systems is shown in Table~\ref{tab:runtime_nc}. We plot the runtime versus the number of GPUs in Figure~\ref{fig:scaling} to show the scaling of each system. 
For these experiments, all systems sample neighbors across the whole graph, and thus reach similar accuracy (e.g., see Table~\ref{tab:runtime_nc} right).

\newparagraph{Key Takeaways}
Across experiments and GPU counts, \systemname is the fastest and cheapest option; runtime and cost reductions are up to 4.5$\times$ and 3.1$\times$ versus existing systems.

\begin{table*}[t]\tiny
    \centering
    \caption{Runtime and cost of DGL, MariusGNN, Salient++, and \systemname on OGBN-Papers100M using a GraphSage-Small (left) and -Large (right) GNN. With disaggregated mini batch preparation, \systemname can scale training from one to eight GPUs while existing systems cannot. Moreover, the extra disaggregated machines are cheap compared to the GPU machines used for model computation and do not prevent reductions in total training cost. \systemname uses 0, 1, 2, and 4 disaggregated batch construction workers for 1-, 2-, 4-, and 8-GPU training respectively. Relative improvement compared to single-GPU training for each system is shown in parentheses.}
    \vspace{0.05in}
    \label{tab:runtime_nc}
    {
        \setlength{\tabcolsep}{5pt}
        \begin{tabular}{l c c c c c c c c c c c c c c c c c}
            \toprule
            & \multicolumn{8}{c}{GraphSage-Small} & \multicolumn{9}{c}{GraphSage-Large} \\
            \cmidrule(lr){2-9}
            \cmidrule(lr){10-18}
            & \multicolumn{4}{c}{Epoch Runtime (s)} & \multicolumn{4}{c}{Epoch Cost (\$)} & \multicolumn{4}{c}{Epoch Runtime (s)} & \multicolumn{4}{c}{Epoch Cost (\$)} & Avg. Acc. \\
            \cmidrule(lr){2-5}
            \cmidrule(lr){6-9}
            \cmidrule(lr){10-13}
            \cmidrule(lr){14-17}
            \cmidrule(lr){18-18}
            \# GPUs & 1 & 2 & 4 & 8 & 1 & 2 & 4 & 8 & 1 & 2 & 4 & 8 & 1 & 2 & 4 & 8 & -\\
            \cmidrule(lr){1-18}
            DGL v2.4 & 143 & 142 & 93.4 & 33.0 (4.3$\times$) & 0.97 & 0.97 & 0.64 & 0.22 (4.4$\times$) & 151 & 146 & 101 & 41.0 (3.7$\times$) & 1.03 & 0.99 & 0.69 & 0.28 (3.6$\times$) & 67.35\\
            MariusGNN & 84.0 & 77.1 & 62.1 & 61.5(1.4$\times$) & 0.57 & 0.52 & 0.42 & 0.42 (1.4$\times$) & 124 & 92.3 & 83.9 & 88.3 (1.4$\times$) & 0.84 & 0.63 & 0.57 & 0.60 (1.4$\times$) & 67.14\\
            Salient++ & 61.5 & 54.5 & 38.1 & 35.6 (1.7$\times$) & 0.42 & 0.37 & 0.26 & 0.24 (1.7$\times$) & 114 & 79.8 & 39.1 & 36.7 (3.1$\times$) & 0.78 & 0.54 & 0.27 & 0.25 (3.1$\times$) & 68.20\\
            \systemname & \textbf{54.9} & \textbf{33.3} & \textbf{14.2} & \textbf{7.35 (7.5$\times$)} & \textbf{0.37} & \textbf{0.25} & \textbf{0.12} & \textbf{0.07 (5.3$\times$)} & \textbf{98.7} & \textbf{50.2} & \textbf{26.1} & \textbf{12.0 (8.2$\times$)} & \textbf{0.67} & \textbf{0.38} & \textbf{0.22} & \textbf{0.12 (5.6$\times$)} & 67.16\\
            \bottomrule
        \end{tabular}
    }
    \vspace{-0.1in}
\end{table*}

\newparagraph{Existing Systems}
We find that existing systems are unable to effectively scale GNN training across multiple GPUs. For GraphSage-Small, the most scalable system (DGL) achieves only a 4.3$\times$ speedup when moving from one to eight GPUs. With a more compute-intensive model (GraphSage-Large), baseline systems are able to scale better---e.g., Salient++ achieves a 3.1$\times$ speedup (rather than 1.7$\times$)---but they still suffer from sublinear speedups as a result of CPU-based mini batch preparation bottlenecks (Section~\ref{sec:prelim}).

\begin{figure}[t]
  \centering
  \includegraphics[width=.45\textwidth]{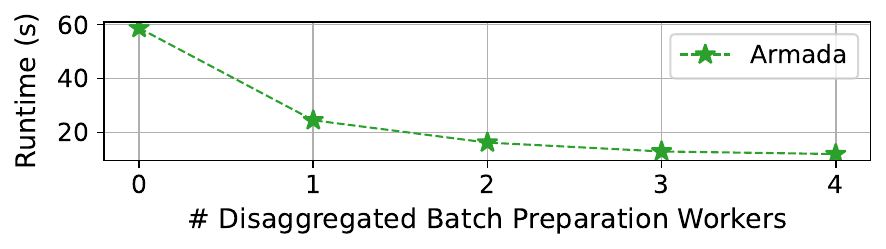}
  \vspace{-0.05in}
  \caption{Epoch runtime in \systemname when training GraphSage-Large on OGBN-Papers100M with eight GPUs and a varying number of disaggregated batch preparation workers; independently scaling these workers allows \systemname to minimize runtime.}
  \label{fig:armada_bw_scaling}
  \vspace{-0.2in}
\end{figure}

\newparagraph{Armada: The Benefit of Disaggregation}
\systemname, however, achieves near-perfect scalability. For GraphSage-Small and -Large respectively, \systemname achieves a 7.5$\times$ and 8$\times$ speedup when moving from one to eight GPUs. The key reason Armada can scale linearly is because of its disaggregated architecture. The effect of disaggregation is evident by comparing \systemname to Armada - Aggregated in Figure~\ref{fig:scaling}. We also show the benefit of disaggregation in Figure~\ref{fig:armada_bw_scaling}; we report the epoch runtime in \systemname when training GraphSage-Large on OGBN-Papers100M with eight GPUs and a varying number of disaggregated batch construction workers. Figure~\ref{fig:armada_bw_scaling} shows that as the number of CPU resources used for mini batch preparation increases, the runtime decreases until the accelerators are fully saturated and the epoch runtime plateaus.

Although the additional machines needed for batch preparation incur additional cost, these machines are cheaper than the GPU machines used for computation. Thus, \systemname is still able to achieve total training cost reductions; we achieve a 5.3$\times$ and 5.6$\times$ reduction in cost when using eight instead of one GPU for GraphSage-Small and -Large respectively. 

\newparagraph{Summary}
Armada's disaggregated architecture allows resource utilization to be optimized in the presence of GNN workload imbalance, leading to linear scaling and cost-effective distributed GNN training over large-scale graphs.

%% file: sections/s7_related_work.tex
\section{Related Work}
\label{sec:related_work}
We highlight related works not previously discussed above.

\newparagraph{Minimizing Mini Batch Preparation Overhead}
To reduce the overhead of mini batch preparation, many prior works seek to offload some of this work to the GPU, either by employing GPU-oriented communication, feature caching, or by directly sampling neighbors on the GPU when possible~\cite{largegcn, min2021pytorchdirect, salient, pagraph, dong2021global, nextdoor}. Other works seek to reduce the overhead of multi-hop neighborhood sampling itself~\cite{graphsage, chen2018fastgcn, zou2019layer, lazygcn, zeng2020graphsaint, clustergcn}. These improvements are orthogonal to our disaggregated architecture and could be incorporated in \systemname. 

\newparagraph{Systems for Large-Scale ML over Graphs}
Many systems have been introduced for training machine learning models over large-scale graphs~\cite{p3gnn, MLSYS2020_ROC, pyg, aligraph, pagraph, dong2021global, gnnsurvey, zheng2020dglke}. Some works focus on scaling training using disk-based storage~\cite{mohoney2021marius, partitioning, pytorchbiggraph}. The most closely related works to \systemname, however, are those that focus on scaling GNN training using distributed multi-GPU or multi-machine settings~\cite{p3gnn, MLSYS2020_ROC, aligraph, wang2021flexgraph}. Like Armada, several of these works aim to reduce cross-machine communication during multi-hop sampling, either by using min-edge-cut partitioning~\cite{distdglv2} or by employing feature replication on each machine~\cite{liu2023bgl, salient++}. To address this challenge, \systemname also supports replicating features and introduces \partitioning to scale min-edge-cut partitioning to large graphs. Among these systems, however, Armada is unique in its use of disaggregation to scale training.

%% file: sections/s8_conclusion.tex
\vspace{0.05in}
\section{Conclusion}
\label{sec:conclusion}
This paper introduced \systemname, a new system for scalable, cost-effective, distributed GNN training. Armada's key contribution is \partitioning, a novel min-edge-cut partitioning algorithm that can efficiently scale to large graphs yet still achieve partition quality comparable to METIS. Armada also introduces a new architecture for GNN training that disaggregates the CPU resources used for GNN neighborhood sampling and feature loading from the GPU resources use for model computation, ensuring that the former can be scaled independently in order to saturate the latter. Overall, our results highlight the promise of new algorithms and systems to democratize large-scale GNN training.